\documentclass[final]{cvpr}

\usepackage{times}
\usepackage{epsfig}
\usepackage{graphicx}
\usepackage{amsmath}
\usepackage{amssymb}
\usepackage{bm}
\usepackage{array}
\usepackage{multirow}
\usepackage{booktabs}
\usepackage{centernot}

\usepackage[ruled,linesnumbered]{algorithm2e}
\newcommand\data[1]{{\normalfont \texttt{#1}}}

\usepackage[pagebackref=true,breaklinks=true,colorlinks,bookmarks=false]{hyperref}

\def\cvprPaperID{1901} 
\def\confYear{CVPR 2023}

\begin{document}

\title{Sharpness-Aware Gradient Matching for Domain Generalization}

\author{{Pengfei Wang$^{1,2}$ \qquad Zhaoxiang Zhang$^{1,2,3}$\thanks{Corresponding author} \qquad Zhen Lei$^{1,2,3}$ \qquad Lei Zhang$^{1\ast}$} \\
	$^1$The Hong Kong Polytechnic University  \quad
  $^2$Center for Artificial Intelligence and Robotics, HKISI, CAS \\
   $^3$ State Key Laboratory of Multimodal Artificial Intelligence Systems, CASIA \\
	{\tt\small \ pengfei.wang@connect.polyu.hk, zhaoxiang.zhang@ia.ac.cn}, \\ {\tt\small zlei@nlpr.ia.ac.cn, cslzhang@comp.polyu.edu.hk}
}


\maketitle

\begin{abstract}
The goal of domain generalization (DG) is to enhance the generalization capability of the model learned from a source domain to other unseen domains. The recently developed Sharpness-Aware Minimization (SAM) method aims to achieve this goal by minimizing the sharpness measure of the loss landscape. 
Though SAM and its variants have demonstrated impressive DG performance, they may not always converge to the desired flat region with a small loss value.
In this paper, we present two conditions to ensure that the model could converge to a flat minimum with a small loss, and present an algorithm, named Sharpness-Aware Gradient Matching (SAGM), to meet the two conditions for improving model generalization capability.
Specifically, the optimization objective of SAGM will simultaneously minimize the empirical risk, the perturbed loss (i.e., the maximum loss within a neighborhood in the parameter space), and the gap between them. By implicitly aligning the gradient directions between the empirical risk and the perturbed loss, SAGM improves the generalization capability over SAM and its variants without increasing the computational cost.
Extensive experimental results show that our proposed SAGM method consistently outperforms the state-of-the-art methods on five DG benchmarks, including PACS, VLCS, OfficeHome, TerraIncognita, and DomainNet.
Codes are available at \url{https://github.com/Wang-pengfei/SAGM}.

\end{abstract}

\section{Introduction} \label{Introduction}
Deep learning methods have achieved remarkable success in various computer vision tasks when the source data and target data are independently and identically distributed (i.i.d). However, the performance of deep models trained in a source domain can drop significantly when applied to unseen target domains. Domain generalization (DG) aims to train a model from a set of source data such that it can be well generalized to new domains without retraining. 
Over the past decade, research on DG has led to a plethora of methods, including those based on domain alignment~\cite{muandet2013icml_DIFL,ganin2016dann,li2018cdann,bahng2019rebias,zhao2020er_entropy_regularization}, meta-learning~\cite{li2018mldg,balaji2018metareg,dou2019masf,zhang2020arm}, and data augmentation~\cite{zhou2021mixstyle,shankar2018crossgrad,carlucci2019jigsaw_jigen}.
Though numerous DG approaches have been proposed, a recent study called DomainBed~\cite{gulrajani2020domainbed} reveals that under a fair evaluation protocol, the naive empirical risk minimization (ERM) method can even outperform most existing DG methods.
Unfortunately, simply minimizing the empirical loss on a complex and non-convex loss landscape is typically insufficient to achieve a good generalization capability~\cite{keskar2016largebatch,garipov2018fge,izmailov2018averaging,foret2020sharpness}. As a result, ERM tends to overfit the training data set and converge to sharp local minima~\cite{foret2020sharpness}.

Recent studies such as Sharpness-Aware Minimization (SAM) \cite{foret2020sharpness} try to improve the model generalization ability by minimizing the sharpness measure of the loss landscape. Denote by $\mathcal{L}(\theta)$ the loss to be minimized (\eg, the cross-entropy loss for classification), where $\theta$ is the parameters of the neural network.
 SAM first adversarially computes a weight perturbation $\epsilon$ that maximizes the empirical risk $\mathcal{L}(\theta)$ and then minimizes the loss of the perturbed network, \ie, $\mathcal{L}(\theta+\epsilon)$. 
Specifically, the objective of SAM is 
to minimize the maximum loss around the model parameter $\theta$. Due to the high complexity of this min-max optimization problem, SAM chooses to minimize an approximation of $\mathcal{L}(\theta)$, denoted by $\mathcal{L}_p(\theta)$ (perturbed loss, see Section \ref{Preliminaries} for details).
However, minimizing $\mathcal{L}_p(\theta)$ is not guaranteed to converge to flat minimum regions \cite{zhuang2022surrogate}.

While $\mathcal{L}_p(\theta)$ may not characterize well the sharpness of loss surface, the surrogate gap $h(\theta) \triangleq \mathcal{L}_p(\theta)- \mathcal{L}(\theta)$ can better describe it. Intuitively, the loss surface will become flatter when $h(\theta)$ is closer to zero because the perturbation parameters around $\theta$ will have a similar loss value. Zhuang \etal \cite{zhuang2022surrogate} proved that $h(\theta)$ is an equivalent measure of the dominant eigenvalue of Hessian (which is the measure of sharpness) at a local minimum. 
Therefore, we can seek flat minima for better generalization ability by minimizing the surrogate gap $h(\theta)$. 
Unlike SAM which only optimizes the perturbation loss $\mathcal{L}_p(\theta)$, GSAM \cite{zhuang2022surrogate} jointly minimizes  $\mathcal{L}_p(\theta)$ and the surrogate gap $h(\theta)$. However, GSAM minimizes 
$h(\theta)$ by increasing the loss $\mathcal{L}(\theta)$, which will reduce the generalization performance. 
Zhang \etal \cite{zhuang2022surrogate} have shown that when the perturbation amplitude $\rho$ is sufficiently small, the surrogate gap is always non-negative, \ie, $\mathcal{L}_p(\theta)\geq\mathcal{L}(\theta), \forall \theta$. Therefore, increasing $\mathcal{L}(\theta)$ will also increase the difficulty of optimizing $\mathcal{L}_p(\theta;\mathcal{D})$.


We propose two conditions that should be met to obtain a model with good generalization performance. (i) First, the loss within a neighborhood of the desired minimum should be sufficiently low. (ii) Second, the minimum is within a flat loss surface. 
More specifically, condition (i) implies a low training loss and represents good performance on the source training data, while condition (ii) reduces the  performance gap of the model on training and testing data. 

Based on the above analysis, we propose a new DG method, namely Sharpness-Aware Gradient Matching (SAGM), to simultaneously minimize three objectives, the empirical risk $\mathcal{L}(\theta)$, the perturbed loss $\mathcal{L}_p(\theta)$ and the surrogate gap $h(\theta)$. By minimizing $\mathcal{L}(\theta)$ and $\mathcal{L}_p(\theta)$, we search for a region with low loss, which satisfies condition (i). By minimizing $h(\theta)$, we avoid steep valleys, which meets condition (ii).
However, optimizing these three objectives simultaneously is difficult due to the inevitable gradient conflicts during training. Fortunately, we find that when the gradient directions of $\mathcal{L}(\theta)$ and $\mathcal{L}_p(\theta)$ are consistent, the gradient direction of $h(\theta)$ is also consistent with them, and hence the gradient descent can be effectively applied to all the three losses.
Therefore, we transform the optimization objective 
into minimizing $\mathcal{L}(\theta)$, $\mathcal{L}_p(\theta)$, and the angle between their gradients, and achieve this goal by implicitly aligning the gradient directions between the $\mathcal{L}(\theta)$ and $\mathcal{L}_p(\theta)$.
The proposed SAGM improves the generalization performance of the model by facilitating the model to converge to a flat region with a small loss
value. Compared with SAM, our SAGM does not increase the computational cost. In addition, SAGM can be combined with previous data augmentation methods, such as Mixstyle \cite{zhou2021domain} for further performance improvements.

The contributions of this work are summarized as follows. First, we analyze the limitations of SAM-like methods and propose two conditions to ensure the model convergence to a flat region with a small loss. Second, we propose the SAGM algorithm to improve the DG capability of deep models. Finally, we demonstrate the superior performance of SAGM to state-of-the-arts on five DG benchmarks.

\section{Related Work}

\textbf{Domain Generalization (DG).}
DG methods aim to learn a model from one or several observed source data sets so that it can be generalized to unseen target domains. 
Existing DG methods handle domain shift from various perspectives, including domain alignment~\cite{muandet2013icml_DIFL,ganin2016dann,li2018cdann,bahng2019rebias,zhao2020er_entropy_regularization}, meta learning ~\cite{li2018mldg,balaji2018metareg,dou2019masf,zhang2020arm}, data augmentation~\cite{zhou2021mixstyle,shankar2018crossgrad,carlucci2019jigsaw_jigen}, disentangled representation learning \cite{peng2019domain,khosla2012undoing,wang2020cross} and capturing causal relation~\cite{arjovsky2019irm,krueger2020vrex}.
Most of the relevant works to this paper address the DG problem from a gradient perspective \cite{li2018mldg,koyama2020out,shi2021gradient,mansilla2021domain}.
For example, Mansilla \etal \cite{mansilla2021domain} proposed a gradient surgery
strategy to tackle DG problems. They designed an AND gate on the inter-domain gradients. The values of the same sign in the gradient are retained, and the values of different signs are set to zero.
While these methods achieve good performance on the training set, they could fail when generalizing to unseen domains because there is no guarantee that the trained models could converge to flat loss regions.

\textbf{Sharpness-Aware Minimization (SAM).}
 SAM \cite{foret2020sharpness} was proposed to solve the sharp minima problem by modifying the objective to minimize a perturbed loss, which is defined as the maximum loss within a neighborhood in the parameter space.
 Liu \etal and Du \etal proposed  LookSAM \cite{liu2022towards} and Efficient SAM (ESAM)~\cite{du2021efficient}, respectively, to address the computation issue of SAM.
These two methods can approach the performance of SAM with less computation, but they inherent the key problem of SAM, \ie, the perturbed loss $\mathcal{L}_p(\theta)$ might disagree with sharpness.
GSAM \cite{zhuang2022surrogate} was then proposed to address this issue. However, GSAM minimizes the surrogate gap $h(\theta) $ at the price of increasing $\mathcal{L}(\theta)$. This limits the potential improvement of generalization capability because increasing $\mathcal{L}(\theta)$ is equivalent to increasing the bound of $\mathcal{L}_p(\theta)$. 
In this paper, we propose a novel SAGM method to facilitate the model converge to a flat region with better generalization ability.

\textbf{Sharpness and Generalization.}
The first study to reveal the relationship between sharpness and model generalization ability can be dated back to ~\cite{1995flat}.
Subsequently, extensive theoretical and empirical studies have been carried out on the relationship between sharpness and generalization from the perspective of loss surface geometry under the i.i.d. assumption~\cite{keskar2016largebatch,dziugaite2017computing,garipov2018fge,izmailov2018averaging,dinh2017sharp,foret2020sharpness}.
For example, Keskar \etal~\cite{keskar2016largebatch} proposed a sharpness measure and revealed the negative correlation between the sharpness measure and the generalization abilities. 
Dinh \etal~\cite{dinh2017sharp} further argued that the sharpness measure can be related to the spectrum of the Hessian, whose eigenvalues encode the curvature information of the loss landscape. 
As for the out-of-distribution (OOD) conditions, a recent work SWAD \cite{cha2021swad} has theoretically shown that a flatter minimum could lead to a smaller DG gap, which provides a theoretical basis for our method. Based on this theoretical finding, Cha \etal \cite{cha2021swad} modified the stochastic weight averaging (SWA) method \cite{izmailov2018averaging}, one of the popular sharpness-aware solvers, by introducing a dense and overfit-aware stochastic weight sampling strategy. Unfortunately, this ensemble-like weighting method could not promote the model to find flatter minima during the training process.

\section{Preliminaries} \label{Preliminaries}
Throughout the paper, we use $\theta$ to denote the weight parameters of a neural network $f$. 
Suppose there are a set of $S$ training source domains, denoted by $\mathcal{D}=\left\{\mathcal{D}_{i}\right\}_{i}^{S}$. 
In the DG setting, usually we assume there is one target domain, denoted by $\mathcal{T}$. 
From each source domain $\mathcal{D}_{i}$, suppose there are $n$ training samples, each consisting of an input $x$ and a target label $y$, \ie, $(x_{j}^{i},y_{j}^{i})_{j=1}^{n}\sim\mathcal{D}_{i}$.
The training loss over all training domains $\mathcal D$ is defined as follows:
\begin{equation}
	\label{L_define}
	\mathcal{L}(\theta;\mathcal{D})=\frac{1}{Sn}\sum_{i=1}^{S}\sum_{j=1}^{n}\ell(f(x^{i}_j;\theta), y^{i}_j)),
\end{equation}
where $\ell$ donates the cross-entropy loss. 
The training of the network is typically a non-convex optimization problem, aiming to search for an optimal weight vector $\hat{\theta}$ that has the lowest empirical risk $\mathcal{L}(\theta;\mathcal{D})$.

The conventional optimization paradigm in Equation \ref{L_define} is called {\em empirical risk minimization} (ERM). In practice, ERM tends to overfit the training set and converge to sharp minima, which can lead to unsatisfactory performance on an unseen domain. The SAM method \cite{foret2020sharpness} is proposed to find a flatter region around the minimum with low loss values. To achieve this goal, SAM solves the following min-max optimization problem:
\begin{equation}
 \min_{\theta} \max_{\epsilon: \|\epsilon\|\leq \rho} \mathcal{L}(\theta + \epsilon;\mathcal{D}), \label{sam}
\end{equation}
where $\rho$ is a predefined constant controlling the radius of the neighborhood. Given $\theta$, the maximization in Equation \ref{sam} finds the weight perturbation $\epsilon$ in the Euclidean ball with radius $\rho$ that maximizes the empirical loss.

Since the maximization in Equation \ref{sam} is generally costly, an approximate maximizer was proposed in \cite{foret2020sharpness} by invoking the Taylor expansion of the empirical loss. 
For a small $\rho$, using Taylor expansion around $\theta$, the inner maximization in Equation~\ref{sam} turns into a linearly constrained optimization with a solution as follows:
 \begin{equation}
 \label{eq:e_define}
  \epsilon = \max_{\epsilon:\Vert\epsilon\Vert \leq \rho}\mathcal{L}(\theta + \epsilon;\mathcal{D}) \approx \rho \frac{\nabla \mathcal{L}(\theta;\mathcal{D})}{\| \nabla \mathcal{L}(\theta;\mathcal{D})\|}.
\end{equation} 
As a result, the optimization problem of SAM reduces to
\begin{equation}
\label{eq:innermax}
 \min_{\theta} \mathcal{L}(\theta + \hat \epsilon;\mathcal{D}), \text{ where } \hat \epsilon \triangleq \rho\frac{\nabla \mathcal{L}(\theta;\mathcal{D})}{\| \nabla \mathcal{L}(\theta;\mathcal{D})\|}.
\end{equation}

For simplicity of notation, the loss function of SAM can be written as the following perturbed loss: 
\begin{equation}
\mathcal{L}_p(\theta;\mathcal{D})=\mathcal{L}(\theta+ \hat \epsilon;\mathcal{D}). 
\end{equation}
In other words, SAM seeks for a minimum on the surface of loss $\mathcal{L}_p(\theta;\mathcal{D})$ rather than  the original loss $\mathcal{L}(\theta;\mathcal{D})$.  

\section{Sharpness-Aware Gradient Matching} \label{SAGM}

In this section, we first analyze the limitations of  SAM-like methods \cite{foret2020sharpness,zhuang2022surrogate} and propose two conditions to ensure the model convergence to a flat loss region. Then we propose a Sharpness-Aware Gradient Matching (SAGM) algorithm that meets these two conditions to achieve good generalization performance. Finally, we provide some analysis to better understand the proposed SAGM method.

\subsection{The perturbed loss $\mathcal{L}_p(\theta;\mathcal{D})$ is not always sharpness-aware}
\begin{figure}
\centering
\includegraphics[width=1.0\linewidth]{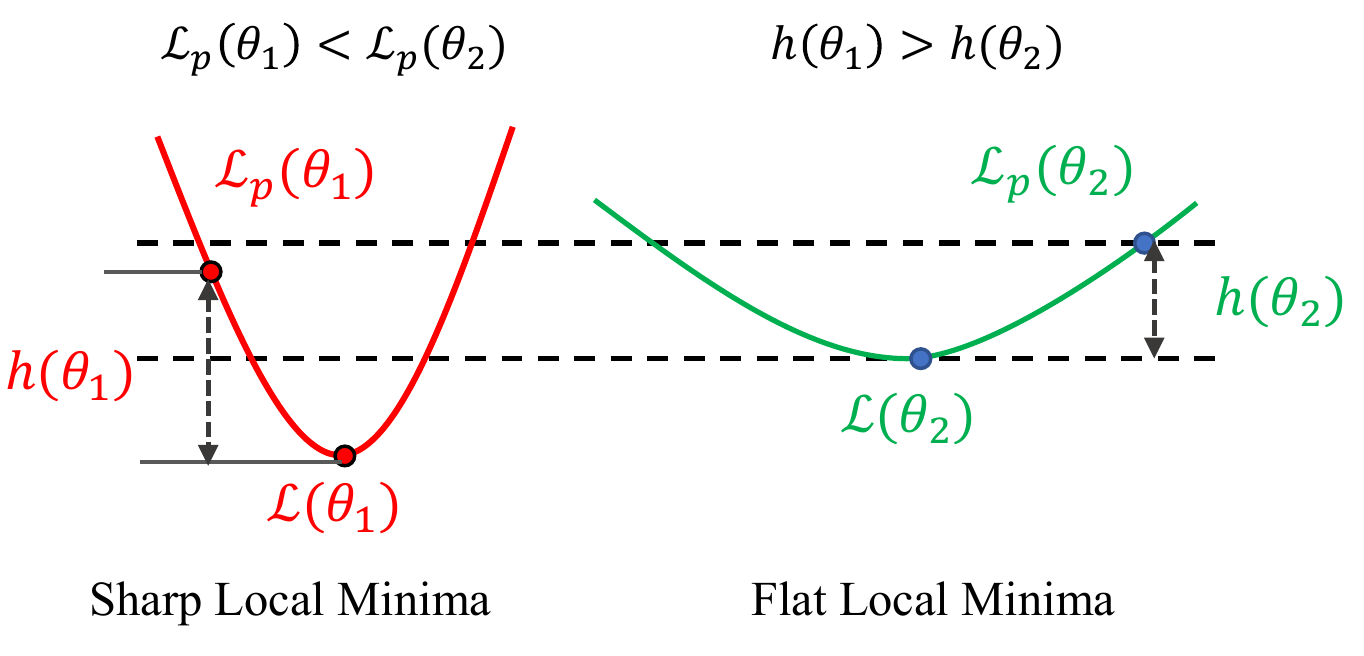}
\caption{
Consider a sharp local minimum $\theta_1$ (red) and a flat local minimum $\theta_2$ (green).  
Since $\mathcal{L}_p(\theta_1;\mathcal{D})<\mathcal{L}_p(\theta_2;\mathcal{D})$ (we omit symbol $\mathcal{D}$ in the figure for simplicity), SAM prefers $\theta_1$ to $\theta_2$. 
However, $\theta_1$ is a sharper minimum than $\theta_2$, and hence $\mathcal{L}_p(\theta;\mathcal{D})$ disagrees with sharpness. Instead, the surrogate gap $h(\theta)$ can better describe the sharpness of loss surface. The smaller value of $h(\theta_2)$ correctly indicates that $\theta_2$ is a flatter minimum than $\theta_1$. Best viewed in color.}
\label{fig: gap_toy}
\end{figure}

Though SAM aims to find a flat region with low loss values, it may not always find such solutions. 
A recent work \cite{zhuang2022surrogate} has proven that for some fixed $\rho$, there is no linear relationship between the perturbed loss and $\sigma_{max}$ (the dominant eigenvalue of the Hessian, a measure of sharpness).
This means that a smaller $\mathcal{L}_p(\theta;\mathcal{D})$ does not guarantee a flatter region. A toy example is shown in Fig.~\ref{fig: gap_toy} to illustrate this. We see that the loss surface of local minimum $\theta_2$ is flatter than that of $\theta_1$. However, SAM will favor $\theta_1$ over $\theta_2$ because $\mathcal{L}_p(\theta_1;\mathcal{D})<\mathcal{L}_p(\theta_2;\mathcal{D})$, resulting in the selection of a sharper minimum.

Fortunately, we find that the surrogate gap, which is defined as:
\begin{equation} \label{eq:gap_def}
    h(\theta) \triangleq \mathcal{L}_p(\theta;\mathcal{D})- \mathcal{L}(\theta;\mathcal{D}),
\end{equation}
can describe the sharpness much better. 
Intuitively, the surrogate gap measures the difference between the maximum loss within the neighborhood and the loss at the minimum point. 
Zhuang \etal  \cite{zhuang2022surrogate} showed that $h(\theta)$ can be used to measure the sharpness. 
Specifically, for a local minimum $\theta^*$, consider the neighborhood centered at $\theta^*$ with a small radius $\rho$, the surrogate gap $h(\theta^*)$ approximately has the following linear relationship with $\sigma_{max}$:
\begin{equation}
    \sigma_{max} \approx 2 h(\theta^*) / \rho^2.
\end{equation}

\begin{figure*}[!t]
\begin{center}
\includegraphics[width=\linewidth]{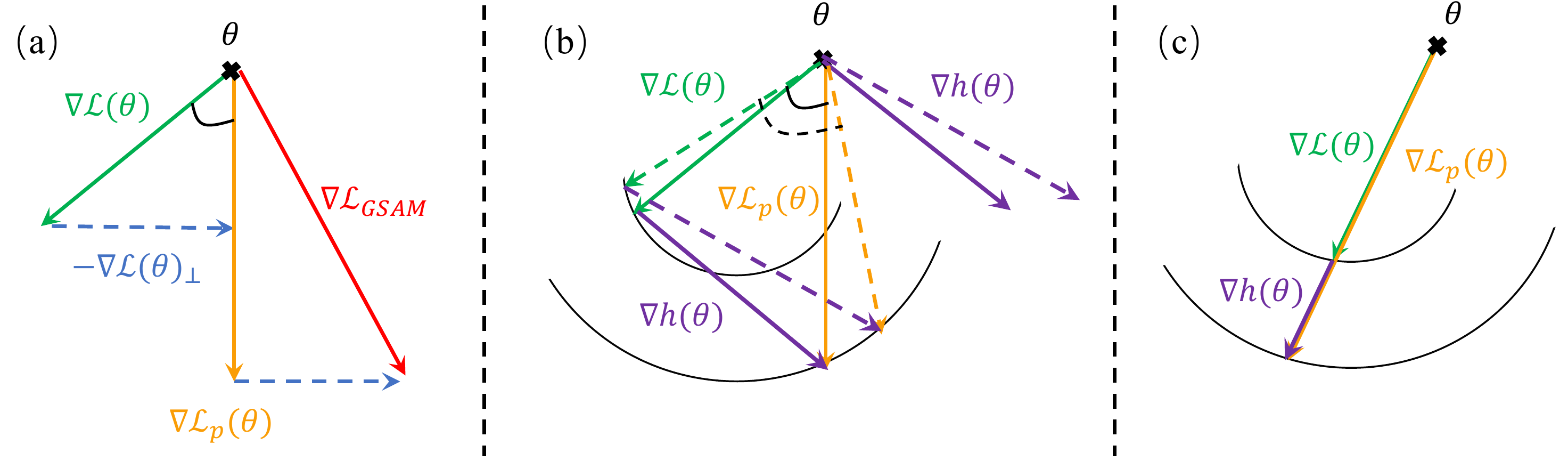}
\end{center}
\caption{
(a) GSAM \cite{zhuang2022surrogate} decomposes $\nabla \mathcal{L}(\theta;\mathcal{D})$ into parallel and vertical components by projecting it onto $\nabla \mathcal{L}_p(\theta;\mathcal{D})$, \ie, ${ \nabla \mathcal{L}_{GSAM}} = { \nabla \mathcal{L}_p(\theta;\mathcal{D})} - \beta { \nabla \mathcal{L}(\theta;\mathcal{D})_\perp}$.
(b) 
When the angle between $\nabla \mathcal{L}(\theta;\mathcal{D})$ and $\nabla \mathcal{L}_p(\theta;\mathcal{D})$ is large, there is an inevitable gradient conflict between $\nabla \mathcal{L}(\theta;\mathcal{D})$, $\nabla \mathcal{L}_p(\theta;\mathcal{D})$, and $\nabla h(\theta)$. 
(c) When the gradient directions of $\nabla \mathcal{L}(\theta;\mathcal{D})$ and $\nabla \mathcal{L}_p(\theta;\mathcal{D})$ are the same, 
$\mathcal{L}(\theta;\mathcal{D})$, 
$\mathcal{L}_p(\theta;\mathcal{D})$ and $h(\theta)$ 
all descend in the most efficient way.
Best viewed in color.
}
\label{fig:method}
\end{figure*}

Based on the above findings, Zhuang \etal \cite{zhuang2022surrogate} proposed the Surrogate Gap Guided Sharpness-Aware Minimization (GSAM) to simultaneously minimize the perturbation loss $\mathcal{L}_p(\theta;\mathcal{D})$ and the surrogate gap $h(\theta)$:
\begin{equation} \label{eq:GAD_objective}
    \min_{\theta}\big(\mathcal{L}_p(\theta;\mathcal{D}),  h(\theta)\big).
\end{equation}
In particular, GSAM decomposes $\nabla \mathcal{L}(\theta;\mathcal{D})$ into two components that are respectively parallel and orthogonal to $\nabla \mathcal{L}_p(\theta;\mathcal{D})$. As shown in Fig. \ref{fig:method} (a), the final gradient direction of GSAM is as follow:
\begin{equation}   \label{eq:obj_GSAM}
\nabla \mathcal{L}_{GSAM} = { \nabla \mathcal{L}_p(\theta;\mathcal{D})} - \beta { \nabla \mathcal{L}(\theta;\mathcal{D})_\perp},
\end{equation}
where $\beta$ is a hyperparameter to scale the stepsize of the ascent step. Note that the horizontal component ${\nabla \mathcal{L}(\theta;\mathcal{D})_\perp}$ keeps $\mathcal{L}_p(\theta;\mathcal{D})$ unchanged because they are perpendicular to each other. However, ${\nabla \mathcal{L}(\theta;\mathcal{D})_\perp}$ will increase $\mathcal{L}(\theta;\mathcal{D})$ 
because there is a gradient conflict between them.
In other words, GSAM reduces the surrogate gap $h(\theta)$ at the price of increasing $\mathcal{L}(\theta;\mathcal{D})$.
We hence argue that GSAM can only lead to sub-optimal generalization ability.  
Since the surrogate gap is always non-negative, the increase of $\mathcal{L}(\theta;\mathcal{D})$ will make the decrease of $\mathcal{L}_p(\theta;\mathcal{D})$  difficult, which may hurt the generalization performance of the model.

\subsection{The objective of SAGM} \label{objective_SAGM}

Inspired by the analysis above, we propose two conditions that should be satisfied to facilitate the model achieving good generalization performance. (i) First, the loss within a neighborhood of the desired minimum should be sufficiently low. (ii) Second, the obtained minimum is within a flat loss surface. 
In order to meet these two conditions, we propose the SAGM method to simultaneously minimize three objectives, the empirical risk loss $\mathcal{L}(\theta;\mathcal{D})$, the perturbed loss $\mathcal{L}_p(\theta;\mathcal{D})$ and the surrogate gap $h(\theta)$:
\begin{equation} \label{aim}
    \min_{\theta}\big(\mathcal{L}(\theta;\mathcal{D}), \mathcal{L}_p(\theta;\mathcal{D}),  h(\theta)\big).
\end{equation}
Intuitively, we search for a region with a low loss by minimizing $\mathcal{L}(\theta;\mathcal{D})$ and $\mathcal{L}_p(\theta;\mathcal{D})$, and we search for a local minimum with a flat surface by minimizing $h(\theta)$. A low loss implies small training errors and represents good performance on the source training data, while a flat loss surface reduces the generalization gap between training and testing performances. When both objectives are reached, the trained model is expected to have high prediction accuracy and good generalization capability.

To achieve our goal, we design a new optimization objective as follows:
\begin{equation}
\min_{\theta} \mathcal{L}(\theta;\mathcal{D}) +  \mathcal{L}_p\big(\theta - \alpha \nabla \mathcal{L}(\theta;\mathcal{D});\mathcal{D}\big), \label{eq:object}
\end{equation}
where $\alpha$ is a hyperparameter. We can rewrite Equation \ref{eq:object} as follows:
\begin{equation}
\min_{\theta} \mathcal{L}(\theta;\mathcal{D}) +  \mathcal{L}\big(\theta + \hat \epsilon - \alpha \nabla \mathcal{L}(\theta;\mathcal{D});\mathcal{D}\big), \label{eq:object2}
\end{equation}
where $\hat \epsilon \triangleq \rho\frac{\nabla \mathcal{L}(\theta;\mathcal{D})}{\| \nabla \mathcal{L}(\theta;\mathcal{D})\|}$ (please refer to Equation \ref{eq:e_define} in Section \ref{Preliminaries} for more details about $\hat \epsilon$). Then we have
\begin{equation}
\min_{\theta} \mathcal{L}(\theta;\mathcal{D}) +  \mathcal{L}\big(\theta + (\frac{\rho}{\vert \vert \nabla \mathcal{L}(\theta;\mathcal{D}) \vert \vert}-\alpha) \nabla \mathcal{L}(\theta;\mathcal{D});\mathcal{D})\big). \label{eq:object4}
\end{equation}

As can be seen from Equation \ref{eq:object4}, our method only needs to calculate losses $\mathcal{L}(\theta;\mathcal{D})$ and $\mathcal{L}\big(\theta + (\frac{\rho}{\vert \vert \nabla \mathcal{L}(\theta;\mathcal{D}) \vert \vert}-\alpha) \nabla \mathcal{L}(\theta;\mathcal{D});\mathcal{D})\big)$. Note that the latter loss has the same computational cost as $\mathcal{L}_p(\theta;\mathcal{D})$.
Considering the fact that $\mathcal{L}(\theta;\mathcal{D})$ has already been calculated when we calculate $\nabla \mathcal{L}(\theta;\mathcal{D})$, SAGM has basically the same computational cost as SAM.

\subsection{Analysis and algorithm of SAGM}
In this section, we provide some analyses to better understand the proposed SAGM method and explain how SAGM satisfies both the two conditions in Section \ref{objective_SAGM}. 

For the second term in Equation \ref{eq:object}, we can perform the first order Taylor expansion around $\theta + \hat \epsilon$ as follows:
\begin{equation}
\begin{aligned}
& \quad\min_{\theta} \mathcal{L}_p\big(\theta - \alpha \nabla \mathcal{L}(\theta;\mathcal{D});\mathcal{D})\big)\\
&=\min_{\theta} \mathcal{L}_p(\theta;\mathcal{D}) - \alpha \nabla\mathcal{L}_p(\theta;\mathcal{D}) \cdot \nabla \mathcal{L}(\theta;\mathcal{D}) + O(\alpha\nabla \mathcal{L}\big(\theta;\mathcal{D})\big)\\
&\approx\min_{\theta} \mathcal{L}_p(\theta;\mathcal{D}) - \alpha \nabla\mathcal{L}_p(\theta;\mathcal{D}) \cdot \nabla \mathcal{L}(\theta;\mathcal{D}).
\label{Taylor expansion}
\end{aligned}
\end{equation}
Therefore, the objective of SAGM in Equation \ref{eq:object} can be rewritten as:
\begin{equation}
\min_{\theta} \mathcal{L}(\theta;\mathcal{D}) + \mathcal{L}_p(\theta;\mathcal{D}) - \alpha \nabla\mathcal{L}_p(\theta;\mathcal{D}) \cdot \nabla \mathcal{L}(\theta;\mathcal{D}). \label{objective_Taylor expansion}
\end{equation}
This equation reveals that we want to \textit{minimize} the loss $\mathcal{L}(\theta;\mathcal{D})$ and $\mathcal{L}_p(\theta;\mathcal{D})$, and \textit{maximize} the inner product of $\nabla\mathcal{L}_p(\theta;\mathcal{D})$ and $\nabla \mathcal{L}(\theta;\mathcal{D})$. 
Minimizing $\mathcal{L}(\theta;\mathcal{D})$ and $\mathcal{L}_p(\theta;\mathcal{D})$ is intuitive and it satisfies condition (i). 
Below we will build the connection between minimizing $h(\theta)$ and maximizing  $\nabla\mathcal{L}_p(\theta;\mathcal{D}) \cdot \nabla \mathcal{L}(\theta;\mathcal{D})$.

First of all, if $\nabla\mathcal{L}_p(\theta;\mathcal{D})$ and $\nabla \mathcal{L}(\theta;\mathcal{D})$ are along similar directions, their inner product is greater than zero, and achieves the peak when their directions are the same.
Therefore, maximizing $\nabla\mathcal{L}_p(\theta;\mathcal{D}) \cdot \nabla \mathcal{L}(\theta;\mathcal{D})$ is actually constraining the directions of these two gradients to be consistent, that is, to perform gradient matching on the gradient of the two losses.
We then investigate the relationship between gradient matching and minimizing $h(\theta)$.
As shown in Fig. \ref{fig:method} (b), 
we draw two circles with $\vert \vert \nabla\mathcal{L}_p(\theta;\mathcal{D})\vert \vert$ and $\vert \vert\nabla \mathcal{L}(\theta;\mathcal{D})\vert \vert$ as radii respectively, and $\nabla h(\theta)$ is the line connecting the respective points on the two circles. Then the gradient direction of $h(\theta)$ is determined by $\nabla\mathcal{L}_p(\theta;\mathcal{D})$ and $\nabla \mathcal{L}(\theta;\mathcal{D})$. 
If the angle between $\nabla\mathcal{L}_p(\theta;\mathcal{D})$ and $\nabla \mathcal{L}(\theta;\mathcal{D})$ is too large, it is hard to optimize $\mathcal{L}(\theta;\mathcal{D})$, $\mathcal{L}_p(\theta;\mathcal{D})$ and $h(\theta)$ simultaneously because of the serious gradient conflict among them.
As shown in Fig. \ref{fig:method} (c), when the gradient directions of $\mathcal{L}(\theta;\mathcal{D})$ and $\mathcal{L}_p(\theta;\mathcal{D})$ are consistent, the gradient direction of $h(\theta)$ will also be consistent with them.
At this time, $\mathcal{L}(\theta;\mathcal{D})$, $\mathcal{L}_p(\theta;\mathcal{D})$ and $h(\theta)$ will descend consistently and efficiently.
Therefore, our optimization objective in Equation \ref{eq:object} can be well achieved by simultaneously minimizing the empirical risk loss $\mathcal{L}(\theta;\mathcal{D})$, the perturbed loss $\mathcal{L}_p(\theta;\mathcal{D})$ and performing gradient matching on
the gradient of the two losses.

The algorithm of SAGM is summarized in \textbf{Algorithm \ref{alg:SAGM}}.



\begin{table*}
\centering
\caption{{\bf Comparison with state-of-the-art domain generalization methods.} Out-of-domain accuracies on five domain generalization benchmarks are shown. The \textbf{best results} are highlighted in bold. 
The results marked by $\dagger, \ddagger$ are copied from Gulrajani and Lopez-Paz \cite{gulrajani2020domainbed} and Cha \etal \cite{cha2021swad}, respectively. Average accuracies and standard errors are calculated from three trials. 
The method ‘Miro (with CLIP)’ uses a pre-trained large-scale CLIP model \cite{radford2021learning} to achieve stronger generalization ability. Following \cite{cha2021swad}, we do not use strategies such as warm up for all methods for a fair comparison.}
\label{table:main_table}
\renewcommand{\arraystretch}{1.1}
\setlength{\tabcolsep}{3pt}
\setlength{\abovetopsep}{0.5em}
\begin{tabular}{l|ccccc|c}
\toprule
\textbf{Algorithm} & \data{PACS}          & \data{VLCS}          & \data{OfficeHome}    & \data{TerraInc}      & \data{DomainNet}     & {Avg.}  \\
\midrule

MMD$^\dagger$ \cite{li2018mmd}                  & 84.7\scriptsize{$\pm0.5$}          & 77.5\scriptsize{$\pm0.9$}          & 66.3\scriptsize{$\pm0.1$}          & 42.2\scriptsize{$\pm1.6$}          & 23.4\scriptsize{$\pm9.5$}          & 58.8           \\
Mixstyle$^\ddagger$ \cite{zhou2021mixstyle}     & 85.2\scriptsize{$\pm0.3$}          & 77.9\scriptsize{$\pm0.5$}          & 60.4\scriptsize{$\pm0.3$}          & 44.0\scriptsize{$\pm0.7$}          & 34.0\scriptsize{$\pm0.1$}          & 60.3           \\
GroupDRO$^\dagger$ \cite{Sagawa2020GroupDRO}    & 84.4\scriptsize{$\pm0.8$}          & 76.7\scriptsize{$\pm0.6$}          & 66.0\scriptsize{$\pm0.7$}          & 43.2\scriptsize{$\pm1.1$}          & 33.3\scriptsize{$\pm0.2$}          & 60.7           \\
IRM$^\dagger$ \cite{arjovsky2019irm}            & 83.5\scriptsize{$\pm0.8$}          & 78.5\scriptsize{$\pm0.5$}          & 64.3\scriptsize{$\pm2.2$}          & 47.6\scriptsize{$\pm0.8$}          & 33.9\scriptsize{$\pm2.8$}          & 61.6           \\
ARM$^\dagger$ \cite{zhang2020arm}               & 85.1\scriptsize{$\pm0.4$}          & 77.6\scriptsize{$\pm0.3$}          & 64.8\scriptsize{$\pm0.3$}          & 45.5\scriptsize{$\pm0.3$}          & 35.5\scriptsize{$\pm0.2$}          & 61.7           \\
VREx$^\dagger$ \cite{krueger2020vrex}           & 84.9\scriptsize{$\pm0.6$}          & 78.3\scriptsize{$\pm0.2$}          & 66.4\scriptsize{$\pm0.6$}          & 46.4\scriptsize{$\pm0.6$}          & 33.6\scriptsize{$\pm2.9$}          & 61.9           \\
CDANN$^\dagger$ \cite{li2018cdann}              & 82.6\scriptsize{$\pm0.9$}          & 77.5\scriptsize{$\pm0.1$}          & 65.8\scriptsize{$\pm1.3$}          & 45.8\scriptsize{$\pm1.6$}          & 38.3\scriptsize{$\pm0.3$}          & 62.0           \\
AND-mask \cite{shahtalebi2021sand} & 84.4\scriptsize{$\pm0.9$} & 78.1\scriptsize{$\pm0.9$}  & 65.6\scriptsize{$\pm0.4$} & 44.6\scriptsize{$\pm0.3$} & 37.2\scriptsize{$\pm0.6$}  &62.0\\
DANN$^\dagger$ \cite{ganin2016dann}             & 83.6\scriptsize{$\pm0.4$}          & 78.6\scriptsize{$\pm0.4$}          & 65.9\scriptsize{$\pm0.6$}          & 46.7\scriptsize{$\pm0.5$}          & 38.3\scriptsize{$\pm0.1$}          & 62.6           \\
RSC$^\dagger$ \cite{huang2020rsc}               & 85.2\scriptsize{$\pm0.9$}          & 77.1\scriptsize{$\pm0.5$}          & 65.5\scriptsize{$\pm0.9$}          & 46.6\scriptsize{$\pm1.0$}          & 38.9\scriptsize{$\pm0.5$}          & 62.7           \\
MTL$^\dagger$ \cite{blanchard2021mtl_marginal_transfer_learning}    & 84.6\scriptsize{$\pm0.5$}          & 77.2\scriptsize{$\pm0.4$}          & 66.4\scriptsize{$\pm0.5$}          & 45.6\scriptsize{$\pm1.2$}          & 40.6\scriptsize{$\pm0.1$}          & 62.9           \\
Mixup$^\dagger$ \cite{xu2020interdomain_mixup_aaai}             & 84.6\scriptsize{$\pm0.6$}          & 77.4\scriptsize{$\pm0.6$}          & 68.1\scriptsize{$\pm0.3$}          & 47.9\scriptsize{$\pm0.8$}          & 39.2\scriptsize{$\pm0.1$}          & 63.4           \\
MLDG$^\dagger$ \cite{li2018mldg}                & 84.9\scriptsize{$\pm1.0$}          & 77.2\scriptsize{$\pm0.4$}          & 66.8\scriptsize{$\pm0.6$}          & 47.7\scriptsize{$\pm0.9$}          & 41.2\scriptsize{$\pm0.1$}          & 63.6           \\
ERM  \cite{vapnik1998statistical}     & 85.5\scriptsize$\pm0.2$ & 77.3\scriptsize$\pm0.4$ & 66.5\scriptsize$\pm0.3$ & 46.1\scriptsize$\pm1.8$ & 43.8\scriptsize$\pm0.1$ & 63.9  \\
Fish \cite{shi2021gradient}                     & 85.5\scriptsize{$\pm0.3$}          & 77.8\scriptsize{$\pm0.3$}          & 68.6\scriptsize{$\pm0.4$}          & 45.1\scriptsize{$\pm1.3$}          & 42.7\scriptsize{$\pm0.2$}          & 63.9           \\ 
SagNet$^\dagger$ \cite{nam2019sagnet}             &{86.3\scriptsize{$\pm0.2$}} & 77.8\scriptsize{$\pm0.5$}          & 68.1\scriptsize{$\pm0.1$}          & 48.6\scriptsize{$\pm1.0$}          & 40.3\scriptsize{$\pm0.1$}          & 64.2           \\
SelfReg \cite{kim2021selfreg}                   & 85.6\scriptsize{$\pm0.4$}          & 77.8\scriptsize{$\pm0.9$}          & 67.9\scriptsize{$\pm0.7$}          & 47.0\scriptsize{$\pm0.3$}          & 42.8\scriptsize{$\pm0.0$}          & 64.2           \\ 
CORAL$^\dagger$ \cite{sun2016coral}             & 86.2\scriptsize{$\pm0.3$}          & 78.8\scriptsize{$\pm0.6$}          & 68.7\scriptsize{$\pm0.3$}          & 47.6\scriptsize{$\pm1.0$}          & 41.5\scriptsize{$\pm0.1$}          & 64.5           \\
mDSDI \cite{bui2021mdsdi}              & 86.2\scriptsize{$\pm0.2$}          & {79.0\scriptsize{$\pm0.3$}}          & 69.2\scriptsize{$\pm0.4$}          & 48.1\scriptsize{$\pm1.4$}          & 42.8\scriptsize{$\pm0.1$}          & 65.1           \\
\midrule
{Miro \cite{cha2022miro} (with CLIP \cite{radford2021learning})}      & 85.4\scriptsize{$\pm0.4$}          & {79.0\scriptsize{$\pm0.0$}} & \textbf{70.5\scriptsize{$\pm0.4$}} & \textbf{50.4\scriptsize{$\pm1.1$}} & {44.3\scriptsize{$\pm0.2$}} & {65.9}  \\
\midrule
SAM $^\ddagger$\cite{foret2020sharpness}  & 85.8\scriptsize$\pm0.2$ & 79.4\scriptsize$\pm0.1$ & 69.6\scriptsize$\pm0.1$ & 43.3\scriptsize$\pm0.7$ & 44.3\scriptsize$\pm0.0$ & 64.5  \\
GSAM \cite{zhuang2022surrogate}       & 85.9\scriptsize$\pm0.1$ & 79.1\scriptsize$\pm0.2$ & 69.3\scriptsize$\pm0.0$ & 47.0\scriptsize$\pm0.8$ & {44.6}\scriptsize$\pm0.2$ & 65.1  \\

\textbf{SAGM (ours)}    & \textbf{86.6\scriptsize{$\pm0.2$}} & \textbf{80.0\scriptsize{$\pm0.3$}}          & {70.1\scriptsize{$\pm0.2$}}       & {48.8\scriptsize{$\pm0.9$}}     & \textbf{45.0\scriptsize{$\pm0.2$}}      & \textbf{66.1} \\

\bottomrule

\end{tabular}
\end{table*}

\begin{algorithm}[h]
    \caption{The Algorithm of SAGM}
    \label{alg:SAGM}

    \KwIn{Domains $\mathcal{D}$; initial weight $\theta_0$; learning rate $\gamma$, batch size, dropout
rate, and weight decay;  hyperparameter $\alpha$ and  radius $\rho$ in Equation~\ref{eq:object4}; total number of iterations $T$.}
    \KwOut{Model trained with SAGM.}
   
    \For{$i \gets 1$ \KwTo $T$}{
     Sample mini-batch $\mathcal{B} \in \mathcal{D}$\;
    Compute the training loss gradient $\nabla \mathcal{L}(\theta_t;\mathcal{B})$\;
        
            Compute $\mathcal{L}_{SAGM}$ according to
 Equation~\ref{eq:object4}: 
$ \mathcal{L}_{SAGM} = \mathcal{L}(\theta_t;\mathcal{B}) +  \mathcal{L}\big(\theta_t + (\frac{\rho}{\vert \vert \nabla \mathcal{L}(\theta_t;\mathcal{B}) \vert \vert}-\alpha) \nabla \mathcal{L}(\theta_t;\mathcal{B});\mathcal{B}\big)$\;
    
   Update weights: \\
   \hspace{5mm} Vanilla  \hspace{1mm} ${\theta}_{t+1} = {\theta}_t - \gamma \nabla \mathcal{L}(\theta_t;\mathcal{B})$ \\
  \hspace{5mm} SAM \hspace{3.5mm} ${\theta}_{t+1} = {\theta}_t - \gamma \nabla \mathcal{L}_p(\theta_t;\mathcal{B})$ \\
  \hspace{5mm} GSAM \hspace{1mm} ${\theta}_{t+1} = {\theta}_t - \gamma \nabla \mathcal{L}_{GSAM}$ (Eq. \ref{eq:obj_GSAM})\\
 \hspace{5mm} SAGM \hspace{1mm} ${\theta}_{t+1} = {\theta}_t - \gamma \nabla \mathcal{L}_{SAGM}$\;

          $i=i+1$\; }
\end{algorithm}

\section{Experiments}
       

\subsection{Experiment setups and implementation details}
\textbf{Dataset and protocol.}
Following \cite{gulrajani2020domainbed}, we exhaustively evaluate our method and competing methods on five benchmarks: {PACS}~\cite{li2017pacs} (9,991 images, 7 classes, and 4 domains), {VLCS}~\cite{fang2013vlcs} (10,729 images, 5 classes, and 4 domains), {OfficeHome}~\cite{venkateswara2017officehome} (15,588 images, 65 classes, and 4 domains), {TerraIncognita}~\cite{beery2018terraincognita} (24,788 images, 10 classes, and 4 domains), and {DomainNet}~\cite{peng2019domainnet} (586,575 images, 345 classes, and 6 domains). 
For a fair comparison, we follow the training and evaluation protocol in DomainBed \cite{gulrajani2020domainbed}, including the dataset splits, hyperparameter (HP) search and model selection on the validation set, except for the HP search space.
Since the original DomainBed setting requires very heavy computation resources, we use a reduced HP search space to reduce computational costs as suggested in \cite{cha2021swad}. 

\textbf{Evaluation protocol.}
In all the experiments, the performances of competing methods are evaluated by the leave-one-out cross-validation scheme. In each case, a single domain is used as the target (test) domain and the others are used as the source (training) domains. The performance is calculated by averaging over three different trials with different train-validation splits. The final accuracy is reported as the average of all possible out-of-domain settings.

\textbf{Implementation details.}
We use ResNet-50 \cite{he2016_cvpr_resnet} pre-trained on ImageNet \cite{russakovsky2015imagenet} as the default backbone. The model is optimized by using the Adam \cite{kingma2015adam} optimizer. 
The hyperparameter $\alpha$ in Equation \ref{eq:object4} is tuned in [0.001, 0.0005].
The hyperparameter $\rho$ is set to 0.05 following SAM \cite{foret2020sharpness}.
The other hyperparameters, such as batch size, learning rate, dropout rate, and weight decay, are tuned in the same search space as that proposed in Cha \etal \cite{cha2021swad}. 
Specifically, we build mini-batches containing examples from all source domains, with 32 examples per domain. The learning rate, dropout rate, and weight decay are tuned in [1e-5, 3e-5, 5e-5], [0.0, 0.1, 0.5], and [1e-4, 1e-6] respectively.
The optimal parameter settings on each dataset are provided in the \textbf{supplementary file}.

\subsection{Main results}
\label{label:s__main_results}
We report the average out-of-domain performances of state-of-the-art DG methods on five benchmarks in Table~\ref{table:main_table}. Due to the limit of space, the accuracies for each domain are reported in the \textbf{supplementary file}. Since SAGM does not rely on domain labels, it can be applied to other robustness tasks that do not contain
domain labels. We also show the generaliztion capability of SAGM on ImageNet and ImageNet-R in the \textbf{supplementary file}.

\textbf{Comparison with conventional DG methods.}
We first compare SAGM with conventional DG methods \cite{li2018mmd,zhou2021mixstyle,Sagawa2020GroupDRO,arjovsky2019irm,zhang2020arm,krueger2020vrex,li2018cdann,shahtalebi2021sand,ganin2016dann,huang2020rsc,blanchard2021mtl_marginal_transfer_learning,xu2020interdomain_mixup_aaai,li2018mldg,shi2021gradient,vapnik1998statistical,nam2019sagnet,kim2021selfreg,sun2016coral,bui2021mdsdi} on the top panel of Table~\ref{table:main_table}. One can see that SAGM consistently outperforms ERM \cite{vapnik1998statistical} on all benchmarks, resulting in +2.2\% average improvement. Meanwhile, SAGM outperforms the leading conventional DG method on each individual dataset: +0.3\% on {PACS} over SagNet \cite{nam2019sagnet} ($86.6\% \rightarrow 86.3\%$); +1.0\% on {VLCS} over mDSDI \cite{bui2021mdsdi} ($79.0\% \rightarrow 80.0\%$); +0.9\% on {OfficeHome} over mDSDI \cite{bui2021mdsdi} ($69.2\% \rightarrow 70.1\%$); +0.7\% on {TerraIncognita} over mDSDI \cite{bui2021mdsdi} ($48.1\% \rightarrow 48.8\%$); and +2.2\% on {DomainNet} over mDSDI \cite{bui2021mdsdi} and SelfReg \cite{kim2021selfreg} ($42.8\% \rightarrow 45.0\%$). 
It is worth mentioning that the performance improvement of SAGM on {DomainNet} is very significant, as {DomainNet} is the largest dataset with 586,575 images from 6 domains.

\textbf{Comparison with Miro (with CLIP) \cite{cha2022miro}.}
SAGM even has an overall performance advantage over Miro, which exploits the large-scale pre-trained CLIP \cite{radford2021learning} model to improve the generalization capability. As shown in Table~\ref{table:main_table}, SAGM outperforms Miro on {PACS}, {VLCS} and {DomainNet}, resulting in +0.2\% average improvement.

\textbf{Comparison with sharpness-aware methods.}
We then compare SAGM with the leading sharpness-aware methods, including SAM \cite{foret2020sharpness} and GSAM \cite{zhuang2022surrogate}.
As shown in the bottom panel of Table~\ref{table:main_table}, SAGM consistently outperforms SAM and GSAM on all the five datasets, leading to an average performance improvement of 1.6\% and 1.0\% over SAM and GSAM, respectively.

Considering the extensive experiments on 5 datasets and 22 target domains, the above results clearly demonstrate the effectiveness and superiority of SAGM in improving the domain generalization capability of deep models.

\subsection{In-domain and out-of-domain performance}
To show that the performance improvement of SAGM is brought by satisfying the two conditions we proposed in Section \ref{objective_SAGM}, we compare SAGM with three classes of generalization methods in more details: data augmentation methods, gradient-based methods, and sharpness-aware methods. In these experiments, we split in-domain datasets into training (60\%), validation (20\%), and test (20\%) splits, and the average results over three runs on PACS are shown in Table~\ref{table:generalization}.
It can be seen that the conventional methods can help in-domain generalization, performing better than ERM on in-domain test set. However, their out-of-domain performances can be worse than ERM.

\textbf{Comparison with data augmentation methods.}
The data augmentation methods, including Mixup \cite{xu2020interdomain_mixup_aaai}, CutMix \cite{yun2019cutmix} and Mixstyle \cite{zhou2021mixstyle}, can improve the in-domain performance over ERM by 0.7\%, 1.0\% and 0.6\%, but their out-of-domain performances are degraded by 0.5\%, 1.5\% and 1.4\%, respectively. 
This is because these data augmentation methods cannot ensure the model to find a flat minima. 
On the contrary, our SAGM method seeks a flat minimum to reduce the performance gap between the source and target domains.
Actually, we can combine SAGM with data augmentation methods such as Mixstyle to further improve the performance. 
Specifically, Mixstyle can transform data $\mathcal{D}$ into $\mathcal{D}_{new}$, which has the same semantics as $\mathcal{D}$ but different styles. 
We can then get a stronger model by minimizing (i) the loss $\mathcal{L}(\theta;\mathcal{D})$ and $\mathcal{L}_p(\theta;\mathcal{D}_{new})$, as well as (ii) the angle between $\nabla\mathcal{L}_p(\theta;\mathcal{D}_{new})$ and $\nabla \mathcal{L}(\theta;\mathcal{D})$. For SAGM+Mixstyle, we insert Mixstyle block after the 1st, 2nd, and 3rd residual blocks following Mixstyle~\cite{zhou2021mixstyle}.
As shown in Table~\ref{table:generalization}, SAGM+Mixstyle obtains further performance improvement over SAGM and Mixstyle.

\begin{table}
\caption{ \textbf{In-domain and out-of-domain performance of different generalization methods on {PACS}.} The scores are averaged over all settings using different target domains. {$\uparrow$} and {$\downarrow$} indicate statistically significant improvement and degradation from ERM.}
\centering
\setlength{\tabcolsep}{3pt}
\begin{tabular}{lcc@{}lcc@{}lcc@{}} 

\toprule
  & \multicolumn{1}{l}{\quad\quad Method}   & \multicolumn{1}{l}{Out-of-domain} & \multicolumn{1}{l}{In-domain}  \\
\midrule
& ERM             & 85.3\scriptsize{$\pm0.4$} & 96.6\scriptsize{$\pm0.0$}                 \\
\midrule

&ERM+Mixup           & 84.8{\scriptsize$\pm0.3$}{$\downarrow$} & 97.3{\scriptsize$\pm0.1$}{$\uparrow$}                      \\
&ERM+CutMix          & 83.8{\scriptsize$\pm0.4$}{$\downarrow$} & 97.6{\scriptsize$\pm0.1$}{$\uparrow$}             \\
&ERM+Mixstyle          & 83.9{\scriptsize$\pm0.3$}{$\downarrow$} & 97.2{\scriptsize$\pm0.2$}{$\uparrow$}             \\
\midrule
&Fish             & 85.3{\scriptsize$\pm0.2$}(-)  & 97.4{\scriptsize$\pm0.3$}{$\uparrow$}                      \\
&AND-mask     & 83.7{\scriptsize$\pm0.5$}{$\downarrow$}  & 97.1{\scriptsize$\pm0.2$}{$\uparrow$}                      \\
&MLDG     & 84.0{\scriptsize$\pm0.2$}{$\downarrow$}  & 97.4{\scriptsize$\pm0.2$}{$\uparrow$}         \\
\midrule
&SAM             & 85.6{\scriptsize$\pm0.1$}{$\uparrow$} & 97.4{\scriptsize$\pm0.1$}{$\uparrow$}                     \\
&GSAM             & 85.8{\scriptsize$\pm0.2$}{$\uparrow$} &97.2{\scriptsize$\pm0.3$}{$\uparrow$}                     \\
\midrule
&SAGM      & 86.2{\scriptsize$\pm0.2$}{$\uparrow$}        & 97.6{\scriptsize$\pm0.1$}{$\uparrow$}                      \\
&SAGM+Mixstyle     & 87.2{\scriptsize$\pm0.2$}{$\uparrow$}        & 97.8{\scriptsize$\pm0.1$}{$\uparrow$}                      \\
\bottomrule
\end{tabular}
\label{table:generalization}
\end{table}

\textbf{Comparison with gradient-based methods.}
From Table~\ref{table:generalization}, we can see that those methods solving the DG problem from the gradient perspective, including Fish \cite{shi2021gradient}, AND-mask \cite{shahtalebi2021sand} and MLDG \cite{li2018mldg}, achieve good in-domain performance, but they fail to improve out-of-domain performance because they cannot guarantee the model converge to a flat region. For example, AND-mask and MLDG improve the in-domain performances by 0.5\% and 0.8\% but degrade the out-of-domain performances by 1.6\% and 1.3\%. 
In contrast, our method achieves better generalization performance under both settings by performing gradient matching of 
$\mathcal{L}(\theta;\mathcal{D})$ and $\mathcal{L}_p(\theta;\mathcal{D})$ during training.


\textbf{Comparison with sharpness-aware methods.}
It can be seen that the sharpness-aware method SAM \cite{foret2020sharpness} can increase both in-domain and out-of-domain performances over ERM, but the improvement on out-of-domain performance is not statistically significant. This suggests that SAM may not find a really flat minimum.
As a comparison, GSAM \cite{zhuang2022surrogate} has a smaller intra-domain and out-of-domain performance gap than SAM.
These experimental results are consistent with our theoretical analysis in Section \ref{SAGM}.
As a comparison, SAGM improves both the in-domain and out-of-domain performances over SAM and GSAM, achieving 0.8\% gain on the out-of-domain performance and 1.0\% gain on the in-domain performance over ERM. 

\begin{table*}[t]
\centering
\small
\caption{{\bf Ablation study on SAGM.} The scores are averaged over all target domain cases. The performances of ERM, SAM, GSAM, ERM+SAM are all optimized by HP searches of DomainBed. 
}
\label{table:ablation}
\renewcommand{\arraystretch}{1.1}
\begin{tabular}{llccccc|c} 
\toprule
           & \data{optimization objective}     & \data{PACS} & \data{VLCS} & \data{OfficeHome} & \data{TerraInc} & \data{DomainNet} & Avg.   \\ \midrule
ERM      &  $\mathcal{L}(\theta;\mathcal{D})$    & 85.5\scriptsize$\pm0.2$ & 77.3\scriptsize$\pm0.4$ & 66.5\scriptsize$\pm0.3$ & 46.1\scriptsize$\pm1.8$ & 43.8\scriptsize$\pm0.1$ & 63.9  \\
SAM     &  $\mathcal{L}_p(\theta;\mathcal{D})$      & 85.8\scriptsize$\pm0.2$ & 79.4\scriptsize$\pm0.1$ & 69.6\scriptsize$\pm0.1$ & 43.3\scriptsize$\pm0.7$ & 44.3\scriptsize$\pm0.0$ & 64.5  \\
GSAM    &  $\mathcal{L}_p(\theta;\mathcal{D}) \downarrow-\mathcal{L}(\theta;\mathcal{D}) \uparrow$       & 85.9\scriptsize$\pm0.1$ & 79.1\scriptsize$\pm0.2$ & 69.3\scriptsize$\pm0.0$ & 47.0\scriptsize$\pm0.8$ & 44.6\scriptsize$\pm0.2$ & 65.1  \\
ERM + SAM    & $\mathcal{L}_p(\theta;\mathcal{D})+\mathcal{L}(\theta;\mathcal{D})$    & 85.6\scriptsize$\pm0.2$ & 78.6\scriptsize$\pm0.2$ & 69.5\scriptsize$\pm0.4$       & 46.4\scriptsize$\pm1.2$     & 44.2\scriptsize$\pm0.0$      & 64.8  \\
SAGM   & $\mathcal{L}(\theta;\mathcal{D}) +  \mathcal{L}_p\big(\theta - \alpha \nabla \mathcal{L}(\theta;\mathcal{D});\mathcal{D}\big)$   & \textbf{86.6\scriptsize{$\pm0.2$}} & \textbf{80.0\scriptsize{$\pm0.3$}}          & \textbf{70.1\scriptsize{$\pm0.2$}}       & \textbf{48.8{$\pm0.4$}}     & \textbf{45.0{$\pm0.1$}}      & \textbf{66.1} \\
\bottomrule
\end{tabular}
\end{table*}

\begin{figure*}[t]
\begin{center}
\setlength{\tabcolsep}{0pt}
\newcommand\figwidth{.255}
\begin{tabular*}{1.0\textwidth}{c @{\extracolsep{\fill}} cccc}

    &
    \raisebox{-0.3\height}{\includegraphics[width=\figwidth\textwidth]{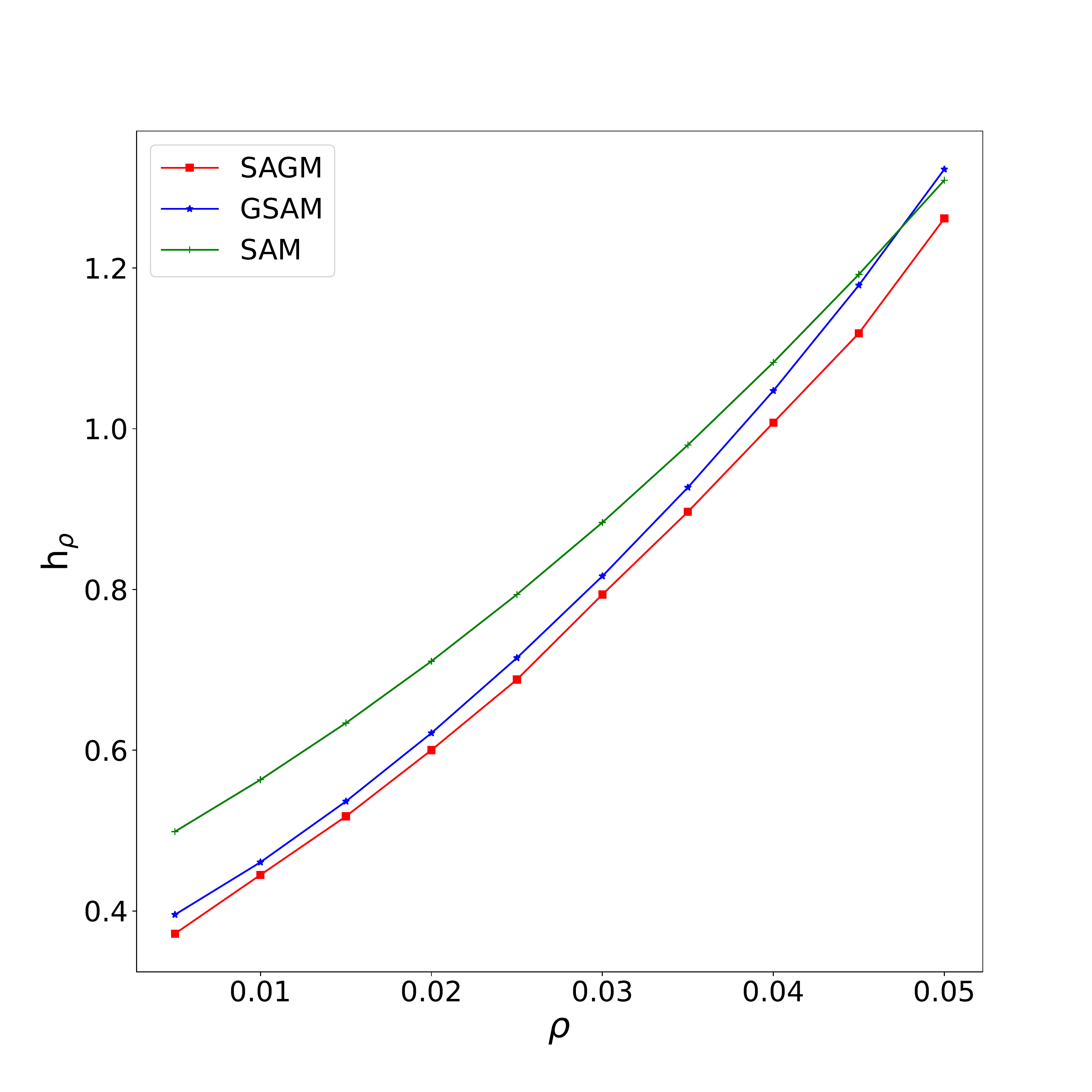}}
    &
    \raisebox{-0.3\height}{\includegraphics[width=\figwidth\textwidth]{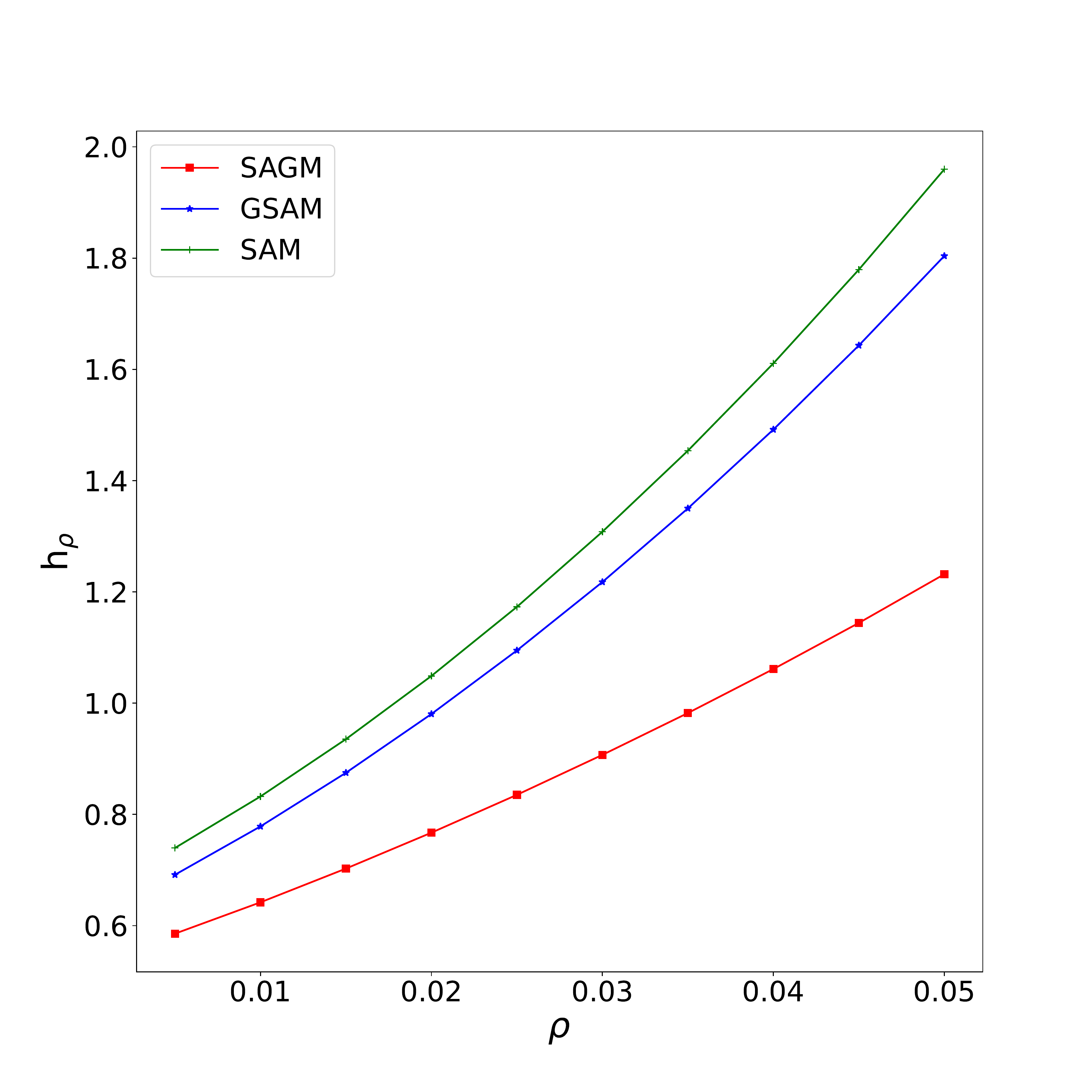}}
      &
    \raisebox{-0.3\height}{\includegraphics[width=\figwidth\textwidth]{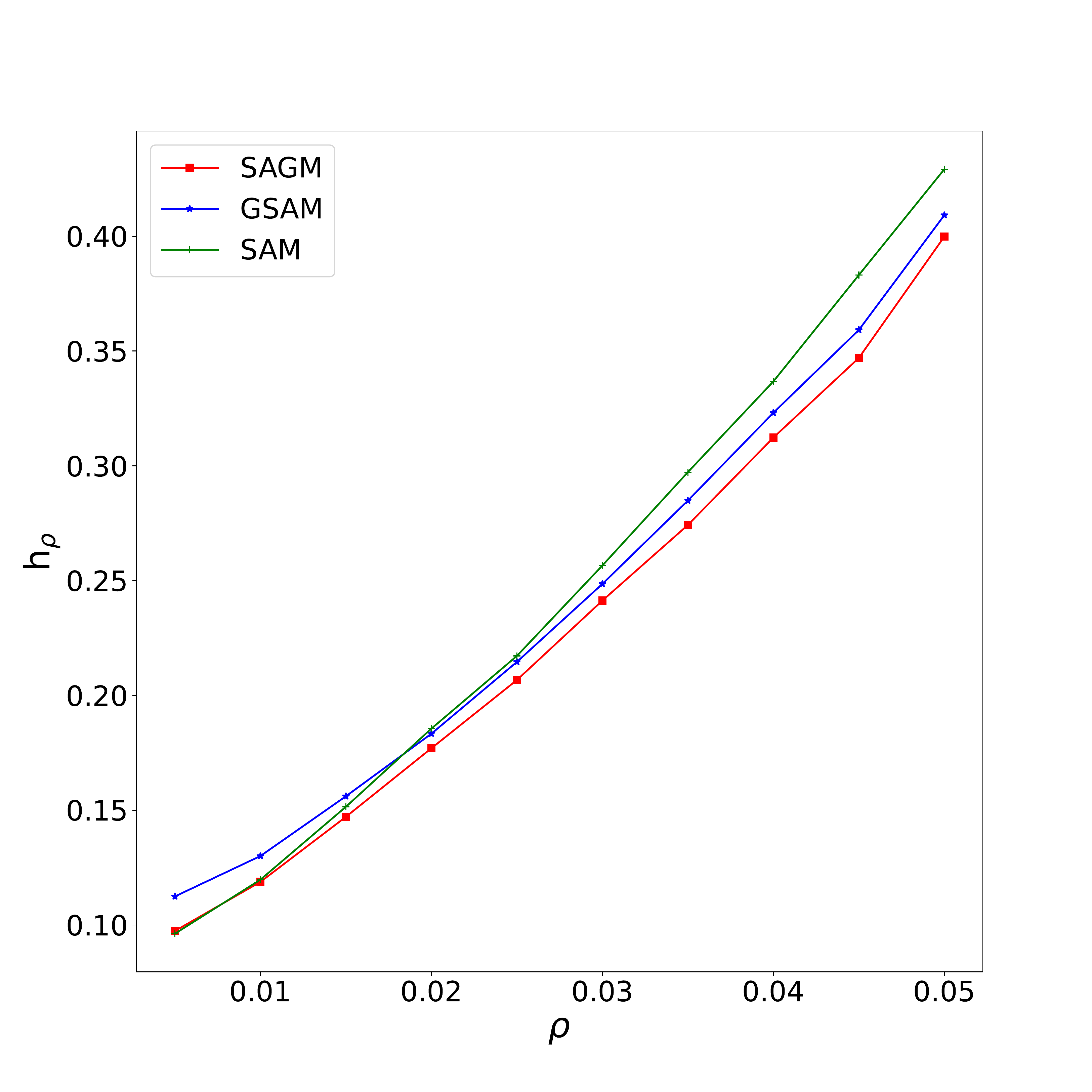}}
    &
    \raisebox{-0.3\height}{\includegraphics[width=\figwidth\textwidth]{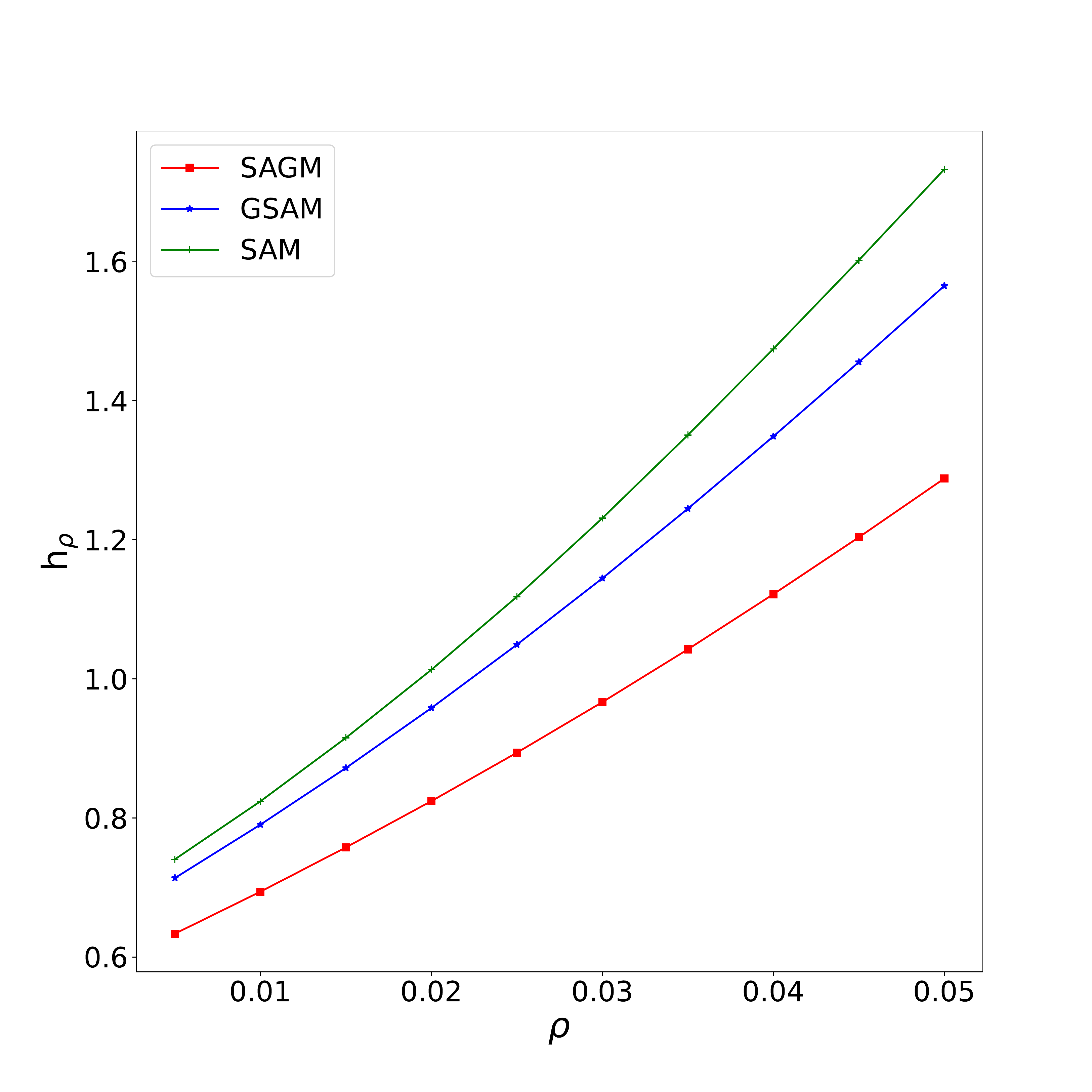}}
\\
    &
    {\small \textit{Art painting} (test)}
    &
    {\small \textit{Cartoon} (test)}
    &
    {\small \textit{Photo} (test)}
    &
    {\small \textit{Sketch} (test)}
\end{tabular*}
\end{center}
\caption{ \textbf{Local sharpness comparison.} We plot the local sharpness via the loss gap of SAM, GSAM and SAGM by varying radius $\rho$ on different target domains of \data{PACS}. For each figure, Y-axis indicates the sharpness ${h}_{\rho}(\theta)$ and X-axis indicates the radius $\rho$. One can see that SAGM finds flatter minima than SAM and GSAM.
    }
\label{fig:sharpness}
\end{figure*}

\subsection{Ablation study}
\label{label:s__ablation}

In the following, we conduct an ablation study to investigate the effectiveness of the optimization objective of SAGM. Specifically, we compare the performance of SAGM with ERM, SAM, GSAM, as well as ERM+SAM, whose optimization objective is $\mathcal{L}_p(\theta;\mathcal{D})+\mathcal{L}(\theta;\mathcal{D})$. 
Compared with SAGM (refer to Equation \ref{objective_Taylor expansion}), ERM+SAM does not have the gradient matching term so that we can use it to investigate the effectiveness of gradient matching in SAGM. Experiments are conducted on the DomainBed benchmark. 

The following observations can be made from Table~\ref{table:ablation}.
First, both SAM, GSAM, and SAGM outperform ERM by significant margins, which verifies the effectiveness of solving the DG problem by seeking a flat minima.
Second, SAGM achieves an average performance improvement of 1.3\% over ERM+SAM, which illustrates the importance of gradient matching in our designed objective function.
Third, SAGM consistently outperforms SAM and GSAM methods by a large margin.
These experimental results verify the effectiveness of the proposed SAGM method.

\subsection{Local sharpness anaylsis}

In this experiment, we show the local sharpness of the trained models to demonstrate that SAGM not only achieves superior DG performance to SAM and GSAM, but also converges to a flatter region in the loss landscape.
To begin with, we quantify the local sharpness of a model parameter $\theta$ by assuming that flat minima will have smaller loss value changes within their neighborhoods than sharp minima.
For the given model parameter $\theta$, we compute the expected loss value changes between $\theta$ and parameters $\theta + \rho\frac{\nabla \mathcal{L}(\theta;\mathcal{D})}{\| \nabla \mathcal{L}(\theta;\mathcal{D})\|}$ with radius $\rho$, \ie, ${h}_{\rho}(\theta)=\mathcal{L}(\theta+ \rho\frac{\nabla \mathcal{L}(\theta;\mathcal{D})}{\| \nabla \mathcal{L}(\theta;\mathcal{D})\|};\mathcal{D})-\mathcal{L}(\theta;\mathcal{D})$. 
Specifically, we compare the ${h}_{\rho}(\theta)$ of SAM, GSAM and our SAGM by varying radius $\rho$ on each target domain of PACS.

From Fig.~\ref{fig:sharpness}, one can see that 
GSAM finds a flatter region than SAM in most experiments, while SAGM finds the flattest minima in all experiments.
It is noted that when the target domain is `Photo (test)', the value of ${h}_{\rho}(\theta)$ is much smaller than other domains. 
This is because SAM, GSAM and our SAGM all achieve over 97.0\% accuracy on this domain (please refer to the \textbf{supplementary file} for the performance on each domain of PACS), which leads to smaller loss value changes. Nonetheless, the trend of `Photo (test)' is consistent with other domains.

\section{Conclusion}
In this paper, we analyzed the limitations of SAM-like algorithms for domain generalization (DG), and proposed two conditions that could ensure the model converges to a flat region with improved generalization ability. We then proposed a novel algorithm named Sharpness-Aware Gradient Matching (SAGM), which minimized simultaneously the empirical loss, perturbed loss and the gap between them, to meet these two conditions.
The proposed SAGM method demonstrated superior performance to state-of-the-art DG methods, including SAM and GSAM, on the five DG benchmarks.
Specifically, SAGM achieved 66.1\% average performance on DomainBed without using any additional information. It even outperformed the Miro \cite{cha2022miro} method that is equipped with the pre-trained CLIP model, demonstrating its strong DG capability.

\noindent\textbf{Acknowledgement.} This work was supported by the InnoHK program.


{\small
\bibliographystyle{ieee_fullname}
\bibliography{cvpr}
}

\end{document}